\documentclass{article}

\usepackage[activate={true,nocompatibility},final,tracking=true,kerning=true,spacing=true,factor=1100,stretch=10,shrink=10]{microtype}

\usepackage{arxiv}

\usepackage[font=small,labelfont={bf}, figurename=Figure, tablename=Table]{caption}

\usepackage[hidelinks]{hyperref}
\hypersetup{
  colorlinks   = true, 
  urlcolor     = esalightblue, 
  linkcolor    = esablue, 
  citecolor   = esablue 
}
\usepackage{amsfonts}       
\usepackage{amsmath}
\usepackage{amssymb}
\usepackage[utf8]{inputenc} 
\usepackage[T1]{fontenc}    
\usepackage{hyperref}       
\usepackage{url}            
\usepackage{booktabs}       
\usepackage{nicefrac}       
\usepackage{lipsum}
\usepackage{graphicx}
\graphicspath{ {./images/} }
\usepackage[nameinlink, capitalise]{cleveref}
\usepackage{xcolor}
\definecolor{esablue}{RGB}{0,57,158}
\definecolor{esalightblue}{RGB}{0,155,219}
\definecolor{esared}{RGB}{150,1,54}
\usepackage{float}

\usepackage{todonotes}

\usepackage[giveninits=true,sorting=none,citestyle=numeric-comp]{biblatex}
\begin{document}

\title{Reconfiguration of pivoting cube ensembles under local sensing constraints using geometric deep learning}

\author{
  Nadezhda Dobreva$^{1}$, Emmanuel Blazquez$^{1}$, Jai Grover$^{1}$, Dario Izzo$^{1}$, Yuzhen Qin$^{2}$, Dominik Dold$^{1,3}$\\[2pt]
  \textit{$^1$ \small{European Space Research and Technology Centre, European Space Agency, Noordwijk, The Netherlands}}\\
  \textit{$^2$ \small{Donders Institute, Radboud University, Nijmegen, The Netherlands}}\\
  \textit{$^3$ \small{Faculty of Mathematics,  University of Vienna, Vienna, Austria}}
}

\maketitle

\begin{abstract}
We demonstrate that local sensing is sufficient for effective global reconfiguration of homogeneous pivoting cube modular robots in two dimensions. While cube selection (i.e., which cube executes a movement) is assumed to be globally coordinated, each cube in the ensemble is controlled by a neural network that only gains information from other cubes in its local neighbourhood, trained using reinforcement learning. Furthermore, we study the effect of including grid symmetries of the cube ensemble (rotation and mirroring) in the neural network architecture. We find that even the most localised versions succeed in reconfiguring to the target shape, although reconfiguration happens faster the more information about the whole ensemble is available to individual cubes. Near-optimal reconfiguration is achieved with only nearest neighbour interactions by using multiple information passing between cubes, allowing them to accumulate more global information about the ensemble. Compared to standard neural network architectures, including grid symmetries provides only minor benefits during training, but allows for reduced model sizes. The presented approach is transferable to other space-relevant systems with different action spaces, such as sliding cube modular robots and CubeSat swarms.
\end{abstract}

\begin{center}
    \large{\textbf{\textsc{Graphical Abstract}}}
\end{center}

\begin{figure}[h!]
    \centering
    \includegraphics[width=.6\linewidth]{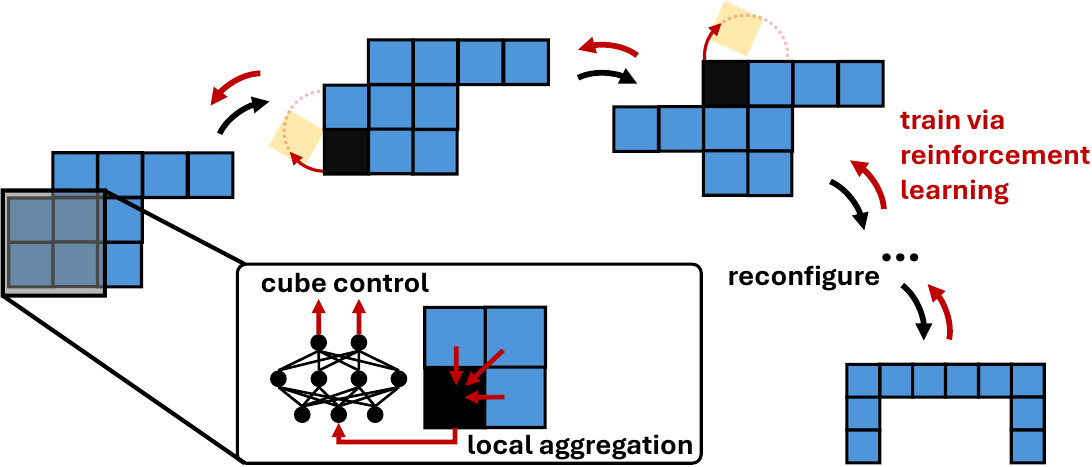}
\end{figure}

We introduce an algorithm for reconfiguring modular cube ensembles under local sensing constraints.
All cubes are identical, equipped with a neural network to control their actuation using locally aggregated information from neighbouring cubes.
Performance is improved by exchanging information with nearest neighbours multiple times, which is sufficient to achieve successful and fast reconfiguration.
Equipping the neural network with inductive biases, i.e., symmetries of the ensemble, only leads to minor improvements.

\keywords{Self-assembly \and Local sensing \and Modular cubes \and Reinforcement Learning \and Symmetries}

\section{Introduction}

Self-assembling and self-configuring structures have the potential to not only reduce the deployment costs of space missions while increasing their life time, but also enable completely new types of missions altogether.
Thus, it comes to no surprise that such concepts are in active development, e.g., for autonomous habitat and space infrastructure construction \cite{ayre2005self,underwood2015using,lee2016architecture,ekblaw2018tesserae,nisser2022electrovoxel,pirat2022toward}, and are already partly used to deploy spacecraft components, such as the primary mirror of the recently launched James Webb Space Telescope (JWST) \cite{rigby2023science}.
Most self-assembling and self-configuring systems follow a common principle: they are composed of several building blocks, such as small robots or spacecraft (or in the case of JWST: mirror elements), which collectively form the desired macroscopic structure.
Assembly and reconfiguration is either guided passively, e.g., quasi-stochastically using electromagnets \cite{hacohen2015meshing,ekblaw2018tesserae}, or actively, which is accomplished using a global or decentralised control scheme \cite{bray2023recent}.

All of these approaches have strengths and weaknesses. Passive guidance is incredibly resource efficient, but comes with the drawback of being slow and mostly limited to assembly scenarios. Global controllers are, arguably, the best approach when it comes to balancing robustness and resource demands. However, they require complete knowledge of the structure's state at all times, thus suffering from increasing communication overhead as the structure grows in size.
Decentralised control, a strategy in which each component uses information available in its immediate local surrounding to control its actuation, offers a middle-ground that avoids the disadvantages of the other two approaches.
However, developing such control schemes comes with challenges of its own, including: (i) Limited processing capability of each decentralised unit which constrains the complexity of models that can be used to guide control. (ii) The information available to guide control is limited, as no single component has view of the entire structure. (iii) A lack of examples of optimal reconfigurations, especially when components are interchangeable, i.e., they can take each other's place in the ensemble. This constrains the type of methods that can be used to, e.g., train local controllers. (iv) The absence of a global clock signal, i.e., the self-organisation process is asynchronous and multiple components can move simultaneously. To enable this kind of movement, mechanisms preventing collisions and invalid configurations are required.

At first glance, global controllers may appear to be the more practical choice. Nevertheless, decentralised control offers unrivaled potential in terms of efficiency and scalability -- which is reflected in the recent focus on decentralised control in robotics \cite{bray2023recent}.
Unsurprisingly, this concept can also be found throughout nature, evident in phenomena ranging from neural ensemble interactions in brain networks to the swarming behavior of animals \cite{o2017oscillators}.
In fact, such decentralised approaches have already been proposed and tested, with a particularly prominent example of decentralised self-organisation being neural cellular automata \cite{mordvintsev2020growing}. There are also applications in the aerospace domain, such as for autonomous mobile robots (e.g., drones and rovers) \cite{de2022insect} and satellite swarms \cite{izzo2005equilibrium,izzo2007autonomous}.

\begin{figure}[t!]
    \centering
    \includegraphics[width=.8\linewidth]{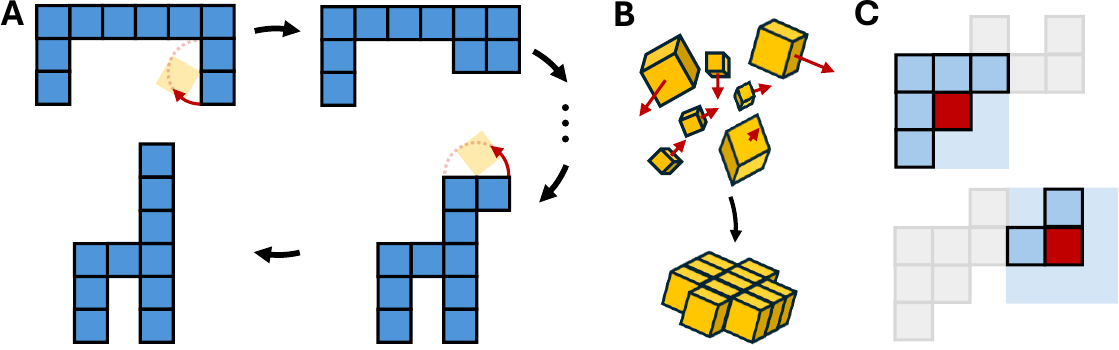}
    \caption{Pivoting cube ensembles enable reconfigurable structures. 
    \textbf{(A)} A cube ensemble in two dimensions reconfigures through pivots between two shapes: a table and a chair. 
    \textbf{(B)} Although the dynamics are more complex for microsatellites such as CubeSats, the reconfiguration problem is structurally similar: all cubes are interchangeable, communication between cubes is locally constrained, and they have to reorganise autonomously to assemble into a target shape (here a mirror surface).
    \textbf{(C)} Only information from cubes in a local neighbourhood is available for a cube to base its actions (i.e., pivots) on, here illustrated for the central red cube.}
    \label{fig:intro}
\end{figure}

In this work, we address the abovementioned challenges (i)-(iii), demonstrating centralised control with local sensing for pivoting cube ensembles, also called ElectroVoxels \cite{nisser2022electrovoxel}.
These ensembles consist of $N \in \mathbb{N}$ cubes connected via their faces.
By correctly activating electromagnets located on the edges of each cube, they are capable of performing pivoting maneuvers relative to one another -- thus enabling the ensemble to reconfigure and reassemble its shape (Fig.~\ref{fig:intro}A).
The reason for choosing this type of system is twofold: first, it features simple update dynamics, allowing us to focus exclusively on developing reconfiguration methodology that can be translated to systems with more complex dynamics, such as Cubesat swarms \cite{pirat2022toward}, see Fig.~\ref{fig:intro}B. Second, ensembles of pivoting cubes are actively being developed \cite{nisser2022electrovoxel}, thus providing a potential future testbed for the introduced methodology.
It has to be emphasised that we do not tackle challenge (iv) though, which concerns asynchronous self-organisation and collision prevention in the absence of a global clock -- leaving fully decentralised control to future work.

Previously, centralised control of pivoting cubes has been realised using an algorithm that morphs the ensemble from any starting to any target shape by first turning it into a line \cite{sung2015reconfiguration}. 
Additionally, control methods have also been trained using deep reinforcement learning. For example, in \cite{song2021autonomous}, a centralised controller observes the global configuration of the ensemble and determines the sequence of actions required to reach the target state.
A related reinforcement-learning-based task-planning strategy has been proposed by \cite{zhang2021reinforcement}, where neural networks are combined with Monte-Carlo Tree Search to compute reconfiguration sequences for modular satellite systems. However, this method also relies on global state information and centralised planning, and assumes labeled, non-interchangeable modules.
In contrast, cubes are identical in our setting, and every cube decides their actions based on locally constrained knowledge of the whole cube ensemble. Thus, we treat the ensemble as a multi-agent system, with every cube controlled by a different instance of the same neural network with local receptive fields.

In our investigated setting, the goal is to reconfigure the ensemble from any initial configuration into a single target configuration -- similar to how neural cellular automata can be trained to always fall into the same pattern \cite{mordvintsev2020growing}.
The two main aspects of our study are locality and interchangeability: every cube is constrained to information in its local environment to inform actions, and all cubes are identical, i.e., they can take each other's place in the ensemble (Fig.~\ref{fig:intro}C).

To achieve this, we map the multi-agent problem to a single-agent problem, which allows us to use standard reinforcement learning libraries to train our agents.
We utilise a convolution-based network architecture that mimics local information aggregation and ensures that all cubes have the same neural network, i.e., are interchangeable, with network parameters trained using Proximal Policy Optimization (PPO) \cite{schulman2017proximal}.
By changing the kernel size or depth of the convolutional neural network, we study the effect of locality on task performance.
Furthermore, we investigate the benefits of using inductive biases of the ensemble's state space in the neural architecture.
For simplicity, we restrict this work to cubes in two dimensions.
We show in experiments that such locally constrained architectures are capable of reconfiguring the cube ensemble, even in the most extreme cases where only nearest neighbour information is available -- although networks with larger receptive fields typically reach the final state much faster.

In the following, we first introduce our methodology and then demonstrate centralised control with local sensing for a variety of assembly scenarios.
The code for all simulations is available on GitHub\footnote{\url{https://github.com/nadiand/CubePy}}.

\section{Methods}

\subsection{Simulating pivoting cube ensembles}

We adapt a custom implementation for simulating two-dimensional pivoting cube ensembles from the \textit{Programmable Cubes Challenge} of the \textit{2024 Space Optimization Competition (SpOC)} \cite{bannach2024space}, which we refer to in the following as the \textit{cube environment}. 
In the cube environment, an ensemble of cubes is represented by a vector of physical coordinates $\pmb r \in \mathbb{R}^{N \times 2}$ (where $N \in \mathbb{N}$ is the number of cubes) and a list of neighbours\footnote{A second list with surrounding cubes (up to a certain radius) is also stored to speed up updates and simplify collision checks.} of each cube $\mathcal{N}_i$, i.e., $\mathcal{N}_i = \{j: \, \,  \| r_i - r_j \|_1 = 1  \}$, with $\| \cdot \|_1$ being the L1 norm and $i \in \{0,1,..., N-1\}$.
\begin{figure}[t!]
    \centering
    \includegraphics[width=1\linewidth]{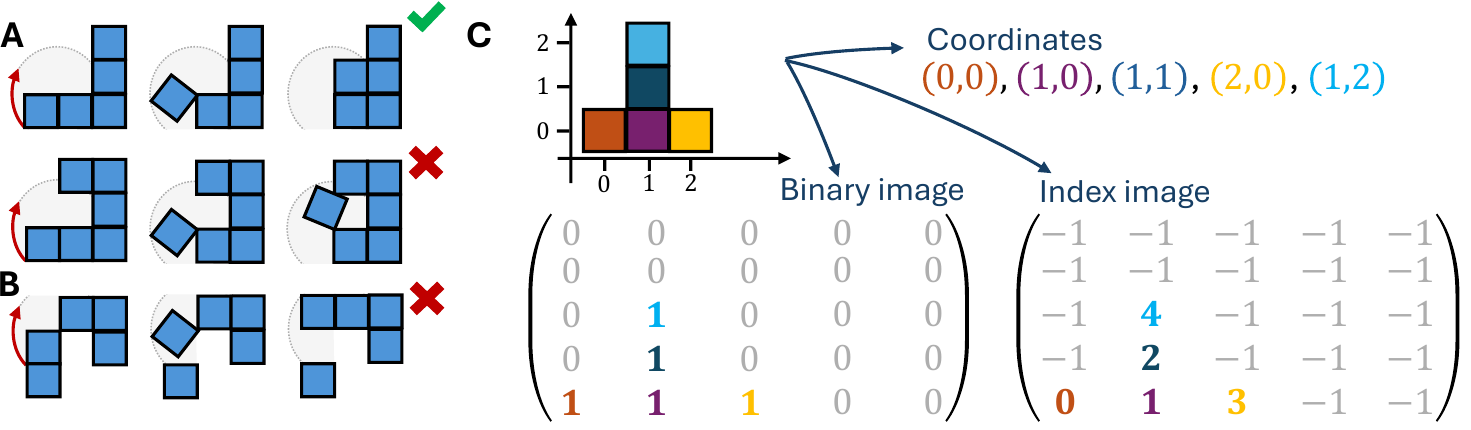}
    \caption{Allowed moves and representation of two-dimensional cube ensembles. 
    \textbf{(A)} Only pivots that do not physically collide with other cubes are allowed. 
    \textbf{(B)} Moreover, the ensemble has to stay connected during reconfiguration. 
    \textbf{(C)} In the cube environment, the ensemble is represented as a list of coordinates. From this representation, we extract two image  representations: a binary image just encoding the shape of the ensemble, and an index image containing the index of each cube in the binary image. The indices correspond to the element in the coordinate list.}
    \label{fig:config}
\end{figure}

For a single update step of the ensemble, both the index of the cube to be moved and the type of move has to be provided.
In the two-dimensional case, there are only two available moves: clock-wise and counter-clock-wise rotation. 
Before updating the ensemble, two checks are performed to ensure that a potential move can be physically realised:
\begin{enumerate}
    \item It is checked whether pivoting the cube would lead to physical collisions with other cubes. This is done by evaluating whether grid locations that would be passed through during the pivot are empty (Fig.~\ref{fig:config}A).
    \item It is checked whether the whole ensemble stays connected after removing the cube to be moved. Or phrased differently: the ensemble is not allowed to fracture into two (or more) disconnected ensembles during reconfiguration (Fig.~\ref{fig:config}B). Therefore, moves that break the connectivity are prohibited.
\end{enumerate}
For the latter, a breadth first search is done on the neighbourhood graph of the ensemble, i.e., cubes are represented as nodes and edges indicate that the faces of two cubes touch.
Assuming cube $k$ is to be moved, we start the search from one of its neighbours $m \in \mathcal{N}_k$ until either (i) all remaining neighbours of cube $k$, $n \in \mathcal{N}_k \backslash \{m\}$, have been visited or (ii) there are no more nodes in the queue that can be visited; with the condition that node $k$ is skipped during the search.
The ensemble only stays connected while moving cube $k$ if all neighbours have been reached during the search.

In the worst case, the whole cube ensemble has to be traversed for the check to terminate, and thus the operation can become fully global.
By constraining the search to only the cubes within a certain local radius of cube $k$, the check can be performed locally instead -- as local connectivity conserves global connectivity as long as the ensemble started from a globally connected configuration.
However, this way, pivots are not guaranteed to be reversible; which is guaranteed when doing the full breadth first search.
As a compromise, we use the full breadth first search during training, while only using a local search -- with the same total receptive field size as the local controllers -- during test time.

\subsection{Controlling the cubes}

\subsubsection{Gym environment}
In our setup, all cubes are identical and interchangeable, which also means that every cube is controlled by the same policy.
Since examples of optimal reconfiguration are not available for supervised learning, we aim at learning policies via reinforcement learning.
To streamline the usage of reinforcement learning methods, we created a Gym \cite{brockman2016openai} environment for the reconfiguring cube scenario which utilises the aforementioned cube environment for simulating cube ensembles. 
To avoid confusion, we will distinguish in the following between the \textit{Gym environment}, which acts as an interface to reinforcement learning methods, and the \textit{cube environment}, which provides the update logic of pivoting cubes and is part of the Gym environment.

The state space of the ensemble is represented as a grid (\textit{binary image}, see Fig.~\ref{fig:config}C), in which an entry of 1 denotes that the cell is occupied by a cube, while an empty cell is denoted by 0. 
We further keep track of the separate cubes and their movements via a corresponding \textit{index image}, which is identical to the binary image but with entries $i \in \{0, 1, ..., N\}$ if the cell is occupied by cube $i$ and -1 if it is empty. 
Thus, the index image contains at all times information about the location of each cube within the ensemble. 
As discussed in the previous section, this information is required for applying updates, i.e., selecting and updating cube coordinates given a move, in the cube environment. 
A visualization of the ensemble representation is given in Fig.~\ref{fig:config}.

The action space is defined by integer numbers $\mathcal{A} = \{0, 1, ..., 2\times N\}$, as each cube has two possible actions to take and only one cube moves at each timestep. Given a predicted action $a \in \mathcal{A}$, we derive the module to move using integer division and the action type (clock-wise vs counter-clock-wise rotation) using the modulo operator:
\begin{align}
    \mathrm{cube\ to\ move} &= \left\lfloor \,\frac{a}{2}\, \right\rfloor \\
    \mathrm{action\ type} &= a \bmod{} 2
\end{align}
Updating the state then works as follows: given an action $a$, we retrieve the cube index $i$ and the action type.
We then update the internal state of the cube environment (i.e., the coordinate list and list of neighbouring cube indices), from which updated versions of the binary and index image are obtained.
If the move turns out to be physically impossible or prohibited, no update is applied.

\subsubsection{Policy network}
\begin{figure}[t!]
    \centering
    \includegraphics[width=1\linewidth]{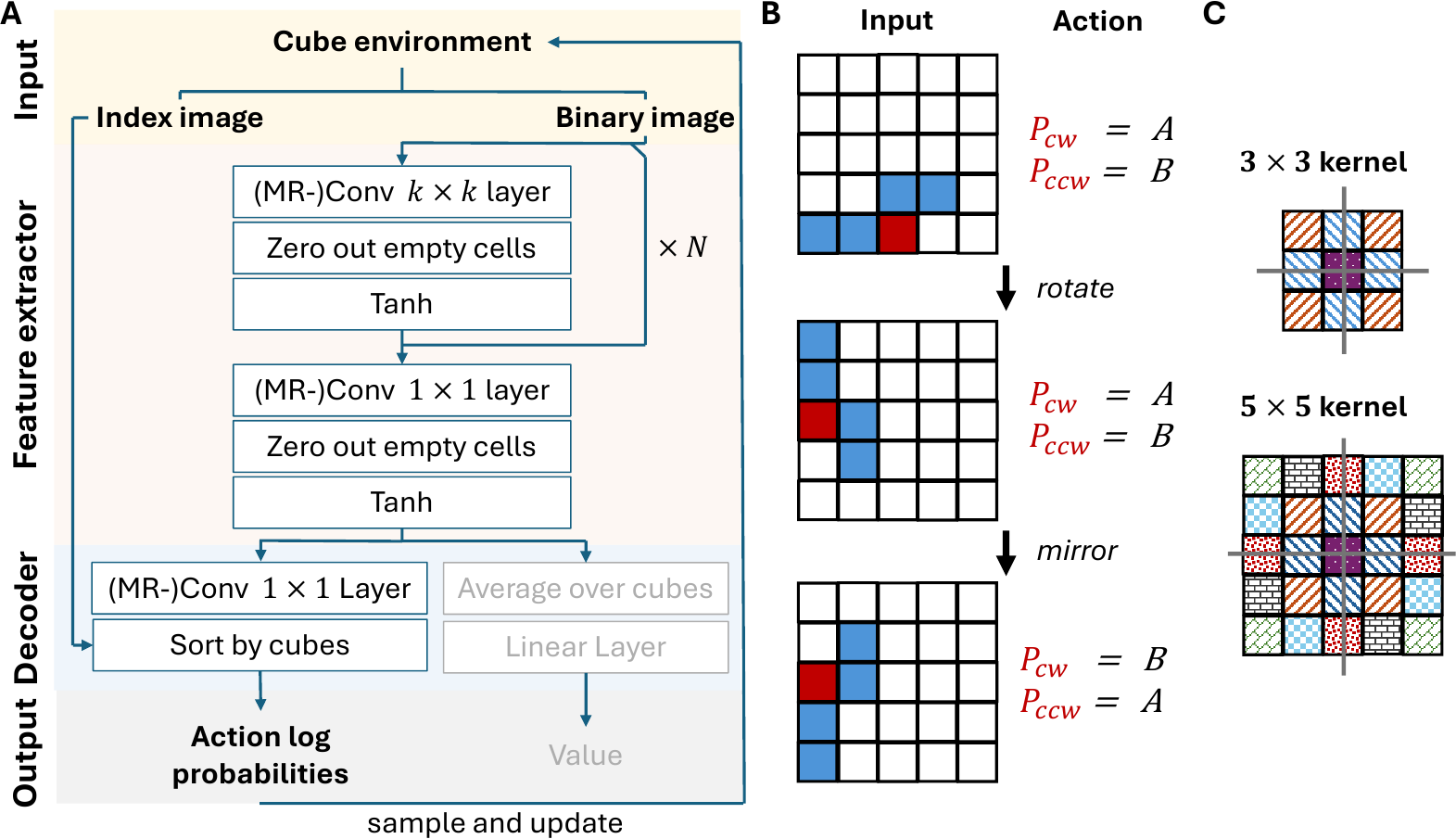}
    \caption{Network architecture. 
    \textbf{(A)} Schematic overview of the model. 
    \textbf{(B)} The two-dimensional grid defining the state space of the cube ensemble has two symmetries.
    If looking at a single cube (red), rotating the ensemble should not change its probabilities for clockwise (cw) and counter-clockwise (ccw) pivoting. Similarly, when mirroring the cube ensemble, these probabilities should simply swap. 
    \textbf{(C)} A $3\times 3$ and $5\times 5$ convolutional kernel.
    Entries with the same colour/pattern share the same value, thus both kernels are invariant under rotation.
    For the $3\times 3$ kernel, rotation invariance also induces mirror invariance (along the gray lines), and thus the alternating behaviour cannot be realised.
    However, for $5\times 5$ kernels, rotation invariance does not enforce mirror invariance.}
    \label{fig:structure}
\end{figure}

The policy that the cubes (or agents) use to pick actions is implemented by a neural network that is semantically divided into a feature extractor and a decoder (Fig.~\ref{fig:structure}A). 
The input of the feature extractor is the binary image described in the previous section, obtained from the cube environment. 
It consists of a number of $k\times k$ convolutional layers, the last one performing a $1\times1$ convolution. 
In-between the convolutions, we mask out the intermediate results using the original binary image input, ensuring that the empty cells are set to 0 before the data is propagated through the next layer. 
This is done to ensure that information is only propagated through cubes and not through empty space.

The decoder has two parts: one decoding the action, and another one decoding the value of the ensemble's current state, which is only used during training. 
To decode the action, first a single $1\times1$ convolution is applied, which maps the output of the feature extractor into 2-channel data: one channel for each possible action, clock-wise and counter-clock-wise rotation. 
Using the index image, this is flattened into a sorted array where element $2k$ and $2k+1$ are the log probabilities of cube $k$ to either pivot clockwise or counterclockwise. 
This array determines the probability distribution of possible actions, used for sampling the next action.
Note that the sampling step is over all cubes, and is hence the only operation in our model that is synchronised and global.
To decode the value, we first average the latent vector over all cubes and apply a single linear layer. 

The convolutional layers realise locally constrained information exchange between neighbouring cubes. The size of the kernel $k$ determines the size of the local neighbourhood (see Fig.~\ref{fig:config}C) -- the larger the kernel size the more global the exchange of information is. Furthermore, the globality increases also the more convolutional layers are added -- even though the individual operations remain local (albeit guided by a global clock signal).
The latter approach can be realised physically, since the $3\times3$ convolutions represent local information exchange between neighbouring, and thus physically connected, cubes.
Therefore, we focus on the approach of stacking convolutional layers with small receptive fields to study the effects of locality on controlling self-assembling cube ensembles. 

\subsection{Mirror-alternating and rotation-invariant convolutions}
The underlying grid structure of the cube ensemble features symmetries under rotation and mirroring. 
For instance, rotating the grid by $90^{\circ}$ should result in the same probability distribution for the possible actions, while mirroring should flip the probabilities, i.e., clockwise rotation becomes counter-clockwise rotation and vice versa (Fig.~\ref{fig:structure}B).
Instead of learning these symmetries from data -- as a normal convolutional layer would do -- they can be integrated into the architecture of the neural network. 
This is, in fact, the idea behind the recently introduced geometric deep learning \cite{bronstein2017geometric}, which allows to increase the expressiveness of a model without increasing its number of parameters.
Most often the number of free parameters is actually reduced due to increased weight sharing. 
In the following, we first describe how rotation invariance is added before introducing the full model.

\subsubsection{Rotation invariance}

Adopting the approach used for \textit{Group Equivariant Convolutional Networks} \cite{cohen2016group}, we introduce rotation-invariant convolutional layers. 
For this aim, we make use of the Reynolds operator \cite{mumford1994geometric}.
First, assume that $G = \{g_0, g_{90}, g_{180}, g_{270}\}$ is the group of rotations by multiples of $90^\circ$.
Furthermore, for $x \in \mathbb{R}^{m}$, $m \in \mathbb{N}$, we have the corresponding matrix representation of the group $\mathcal{R} = \{R_0, R_{90}, R_{180}, R_{270}\}$ with $r \in \mathbb{R}^{m \times m}$ for $r \in \mathcal{R}$, i.e., $R_{90}x$ rotates $x$ by $90^\circ$.
The Reynolds operator is then given by
\begin{equation}
    \bar{R} = \frac{1}{4} \sum_{r \in \mathcal{R}} r \,.
\end{equation}
By definition, this operator is invariant under rotation by multiples of $90^\circ$, meaning that for any $r \in \mathcal{R}$, we have $\bar{R} (r x) = \bar{R} x$.
Since $\bar{R}$ is a projection operator, all its eigenvalues are either $0$ or $1$.
To construct a general weight matrix $W$ which is invariant to rotated inputs, we thus use the left eigenspace of $\bar{R}$ corresponding to the eigenvalue $1$, with basis vectors $B_i$ \cite{mouli2021neural}:
\begin{equation}
    W = \sum_i \omega_i B_i \,.
\end{equation}
The resulting weights are convolved with the input to produce the output of the network layer. 
During training, only the coefficients of the basis functions $\omega_i$ are learned, guaranteeing that $W$ is always rotation invariant.

\subsubsection{Alternating mirror symmetry}

The mirror symmetry is built into the model using the rotation-invariant convolutions.
First, we introduce the horizontal ($\mathcal{M}_H$) and vertical ($\mathcal{M}_V$) mirror operator, i.e., $\mathcal{M}_H(A)_{ij} = A_{ik}$ with $k = m-j$ and $A \in \mathbb{R}^{m \times m}$, $m \in \mathbb{N}$.

The first convolutional layer is special, as it splits the input into two channels $x^{(0)}$ and $y^{(0)}$ which swap places when the cube ensemble is mirrored. Let $W \in \mathbb{R}^{k \times k}$ be a rotation-invariant matrix and $b$ a bias term. 
Then $x^{(0)}$ and $y^{(0)}$ are given by:
\begin{align}
    x^{(0)} = W * x + b \,,\\
    y^{(0)} = \mathcal{M}_H(W) * x + b \,,
\end{align}
where $*$ is the convolution operator.
The next layer receives both $x^{(0)}$ and $y^{(0)}$ as input and consists of two rotation-invariant matrices $W_0^{(1)} \in \mathbb{R}^{k \times k}$ and $W_1^{(1)} \in \mathbb{R}^{k \times k}$:
\begin{align}
    x^{(1)} = \frac{1}{2} \left( W_0^{(1)} * x^{(0)} +  W_1^{(1)} * y^{(0)}\right) + b^{(1)} \,,\\
    y^{(1)} = \frac{1}{2} \left( W_1^{(1)} * x^{(0)} +  W_0^{(1)} * y^{(0)}\right) + b^{(1)} \,.
\end{align}
Again, if $x^{(0)}$ and $y^{(0)}$ swap places, $x^{(1)}$ and $y^{(1)}$ do so as well -- realising the desired alternating behaviour. 
Assuming $N$ convolutions, the subsequent $1 \times 1$ convolution is given by:
\begin{align}
    x^{(N)} = \frac{1}{2} \left( W_0^{(N)} * x^{(N-1)} +  W_1^{(N)} * y^{(N-1)}\right) + b^{(N)}\\
    y^{(N)} = \frac{1}{2} \left( W_1^{(N)} * x^{(N-1)} +  W_0^{(N)} * y^{(N-1)}\right) + b^{(N)}
\end{align}
where $W_0^{(2)}$ and $W_1^{(2)}$ are ordinary weight matrices.
Finally, a last $1 \times 1$ convolution is applied:
\begin{align}
    o_0 = \frac{1}{2} \left( R_0 * x^{(n)} +  R_1 * y^{(n)}\right) + b\\
    o_1 = \frac{1}{2} \left( R_1 * x^{(n)} +  R_0 * y^{(n)}\right) + b
\end{align}
where $o_0$ contains the log probabilities for each cube to pivot clockwise, and $o_1$ to pivot counter-clockwise.
$R_0$ and $R_1$ are ordinary weight matrices.
The input to the value network is simply the average over both channels, $\frac{1}{2}(x^{(N)} + y^{(N)})$.

For $k=3$, the rotation-invariance does not allow mirror symmetry (Fig.~\ref{fig:structure}C).
Thus, in this case we only build networks with rotation-invariant weight matrices.
Starting with $k=5$, both rotation and mirror symmetries are built into the convolution operators.
We denote these layers as Mirror-Rotation-Invariant Convolution (\textit{MR-CNN}) layers.

\subsection{Training networks using reinforcement learning}
The policy network is trained using the policy gradient method PPO. 
We use the \texttt{stable-baselines3} framework, which provides implementations of commonly used reinforcement learning algorithms, such as PPO, that can be utilised almost out-of-the-box.
We choose PPO in particular because of its relative ease of use, and its rather fast training in comparison to considered alternatives, such as Deep Q-Learning \cite{kozlica2023deep}. 
In addition to the policy network, PPO includes a value network which estimates the cumulative value of a given state. 
This network's evaluation contributes to one of the loss terms being optimised during training and is not used at inference time.
The value network often shares architecture with the policy network, as is the case in our architecture design (Fig.~\ref{fig:structure}A). 

We do not use the standard implementation of PPO as we include action masking\footnote{This is facilitated via the \texttt{stable-baselines3-contrib} library, which contains experimental reinforcement learning algorithms and tools, before their inclusion in the official framework. Masked PPO and action masking are included in the library.}. 
Masking out illegal actions by setting their log probabilities to $-\infty$ prevents the policy network from learning to make such undesirable moves.
During training, mask values are obtained using global information (i.e., a full breadth first search), while during testing, only local information is used for masking the actions of a cube. This reflects the usual development phases: training happens in a supervised environment, while testing imitates the condition after deployment.

Empirically, we found that the following reward $r(t)$ at step $t$ works well for our environment:
\begin{equation}
r(t) =
\begin{cases} 
1, & \text{if } O(t) = O_{\text{max}} \,,\\[10pt]
\frac{\alpha_0}{S_\text{max}} \left( \frac{O(t)}{O_{\text{max}}} \right)^{\gamma_0}, & \text{else if } O(t) \geq O(t-1)\,, \\[10pt]
-\frac{\alpha_1}{S_\text{max}} \left( \frac{O(t)}{O_{\text{max}}} \right)^{\gamma_1}, & \text{otherwise}\,,
\end{cases}
\end{equation}
where $S_\text{max}$ is the number of maximum steps per episode, $O(t)$ the overlap of the ensemble with the target shape at step $t$ (i.e., how many blocks are correctly in place), and $O_\text{max}$ the maximum overlap (the number of cubes).
In simulations, we use $\alpha_0 = \alpha_1 = 0.7$ and $\gamma_0 = \gamma_1 = 1.2$, obtained from a parameter sweep.

The overlap $O(t)$ is calculated using phase correlation, an approach for estimating the offset between two images using their frequency-domain representation, commonly used in image registration \cite{kuglin1975phase}:
Given the symmetry of the grid both for rotation and mirroring, we compare the current configuration to the target after applying all possible transformations, i.e., 8 comparisons in total. 
Thus, the problem becomes a matter of translation between the target and each of the 8 configurations, and for that we can compute the cross-correlation between each pair, by means of fast Fourier transform and locating its peak \cite{guizar2008efficient}.
$O(t)$ is then obtained from the result that produces maximum overlap with the target shape.

\subsection{Experimental details}\label{sec:exps}

\noindent We investigate different models using the following design parameters:
\begin{itemize}
    \item \textbf{Network type:} CNN, MR-CNN.
    \item \textbf{Kernel size:} $3\times 3, 5\times 5$.
    \item \textbf{Number of convolutional layers:} $1 -- 4$. Depending on this parameter, there are different number of neurons per layer: $[1, 8192, 32]$ for a single-layer network, $[1, 64, 512, 32]$ for a 2-layer network, $[1, 64, 256, 128, 32]$ for a 3-layer network, and $[1, 64, 256, 128, 64, 32]$ for a 4-layer network.
\end{itemize}
The steps taken per episode (and in total) in PPO depend on the target shape (\textbf{line:} 300 steps per episode, 40000 in total; \textbf{table and chair}: 800, 100000; \textbf{sun-shield:} 2000, 120000). 
For the results reported in Fig.~\ref{fig:training} and Fig.~\ref{fig:rewards}, experiments are repeated five times with different random seeds. 
For the results shown in Fig~\ref{fig:repairing}, we show $14$ repeats with different random seeds.

In Table~\ref{tab:architecture_summary}, we summarise the network architectures used throughout this study, with estimates for the memory requirements of the models (based on saving parameters as floats). The used architectures are small in size, comparable to neural networks used for other onboard applications.
For example, for asteroid geodesy, deep neural networks with $9$ layers of $100$ neurons have recently been proposed for onboard application \cite{izzo2022geodesy}, while for spacecraft guidance \& control, 4-layer neural networks with 700 neurons per layer have been used \cite{izzo2023neural} -- although recent work reports strongly reduced network sizes, with only 37,636 parameters \cite{origer2026guidance}.
Similarly, during ESA's OPS-SAT challenge \cite{meoni2024ops}, a model size of $10$MB was imposed for onboard deployment, which our models satisfy.
Regarding model size, the MR-CNN architectures feature lower memory requirements than standard CNNs due to the increase in shared parameters.

\begin{table}
\caption{Summary of the evaluated network architectures and their model size.}
\label{tab:architecture_summary}
\centering
\setlength{\tabcolsep}{5pt}
\begin{tabular}{lllll}
\toprule
Model type & Kernel & Channel layout & \# Parameters & Save size (float32) \\
\midrule
CNN & 3x3 & $[1, 8192, 32]$ & 344,195 & 1.31 MiB \\
CNN & 3x3 & $[1, 64, 512, 32]$ & 312,579 & 1.19 MiB \\
CNN & 3x3 & $[1, 64, 256, 128, 32]$ & 447,619 & 1.71 MiB \\
CNN & 3x3 & $[1, 64, 128, 128, 64, 32]$ & 298,051 & 1.14 MiB \\
CNN & 3x3 & $[1, 64, 256, 128, 64, 32]$ & 519,363 & 1.98 MiB \\
MR-CNN & 3x3 & $[1, 8192, 32]$ & 295,043 & 1.13 MiB \\
MR-CNN & 3x3 & $[1, 64, 512, 32]$ & 115,587 & 0.44 MiB \\
MR-CNN & 3x3 & $[1, 64, 256, 128, 32]$ & 152,323 & 0.58 MiB \\
MR-CNN & 3x3 & $[1, 64, 256, 128, 64, 32]$ & 174,915 & 0.67 MiB \\
MR-CNN & 5x5 & $[1, 8192, 32]$ & 589,987 & 2.25 MiB \\
MR-CNN & 5x5 & $[1, 64, 512, 32]$ & 492,707 & 1.88 MiB \\
\bottomrule
\end{tabular}
\end{table}

\section{Results}

We assess the efficacy of locally constrained control by training ensembles of squares for four target shapes: a line, table, and chair made of nine cubes, and a ``sun-shield'' made of eleven cubes (Fig.~\ref{fig:training}, top). 
Due to the high variance generally observed in reinforcement learning algorithms \cite{henderson2018deep} and to account for sensitivity to initial conditions, performance for every setting is presented over 5 trials using different random seeds. 
Evaluation of the trained networks is conducted using 500 random initial cube ensemble configurations. 
Two metrics are recorded: success rate, i.e., the percentage of successful reconfigurations from a random initial state to the target shape, and number of moves, i.e., the number of pivots needed to reach the target configuration if reconfiguration is successful.

\subsection{The effect of locality on reconfiguration success}

\begin{figure}[b!]
    \centering
    \includegraphics[width=\linewidth]{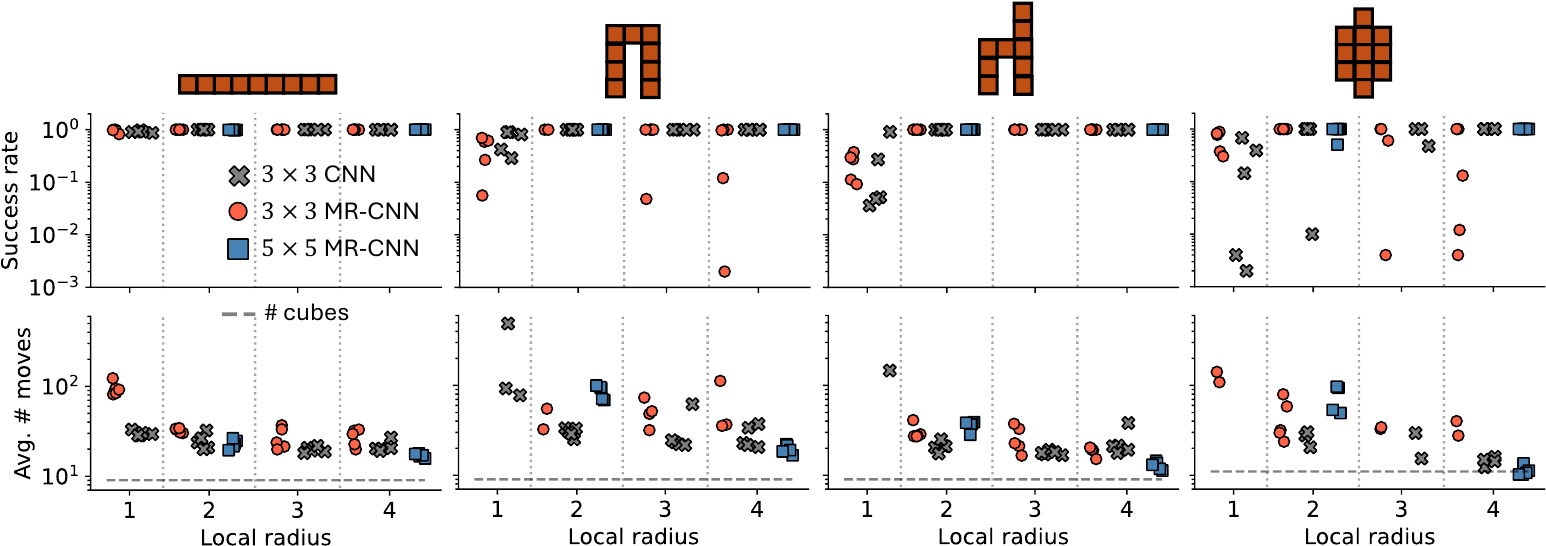}
    \caption{Performance for different target shapes and network architectures. 
    We show the success rate for ensembles reconfiguring into different target shapes from any random initial shape, both using ordinary CNNs and MR-CNNs. 
    The local radius indicates from how far away cubes accumulate information to choose their action.
    For instance, a one layer network with kernel $3\times 3$ has a local radius of $1$, two layers with kernel $3 \times 3$ a radius of $2$, and a one layer network with kernel $5\times 5$ a radius of 2.
    Furthermore, the average number of moves required to reach the target is shown for networks with a success rate exceeding $80$\%.
    Simulations were repeated five times for different random seeds.}
    \label{fig:training}
\end{figure}

\begin{figure}[t!]
    \centering
    \includegraphics[width=\linewidth]{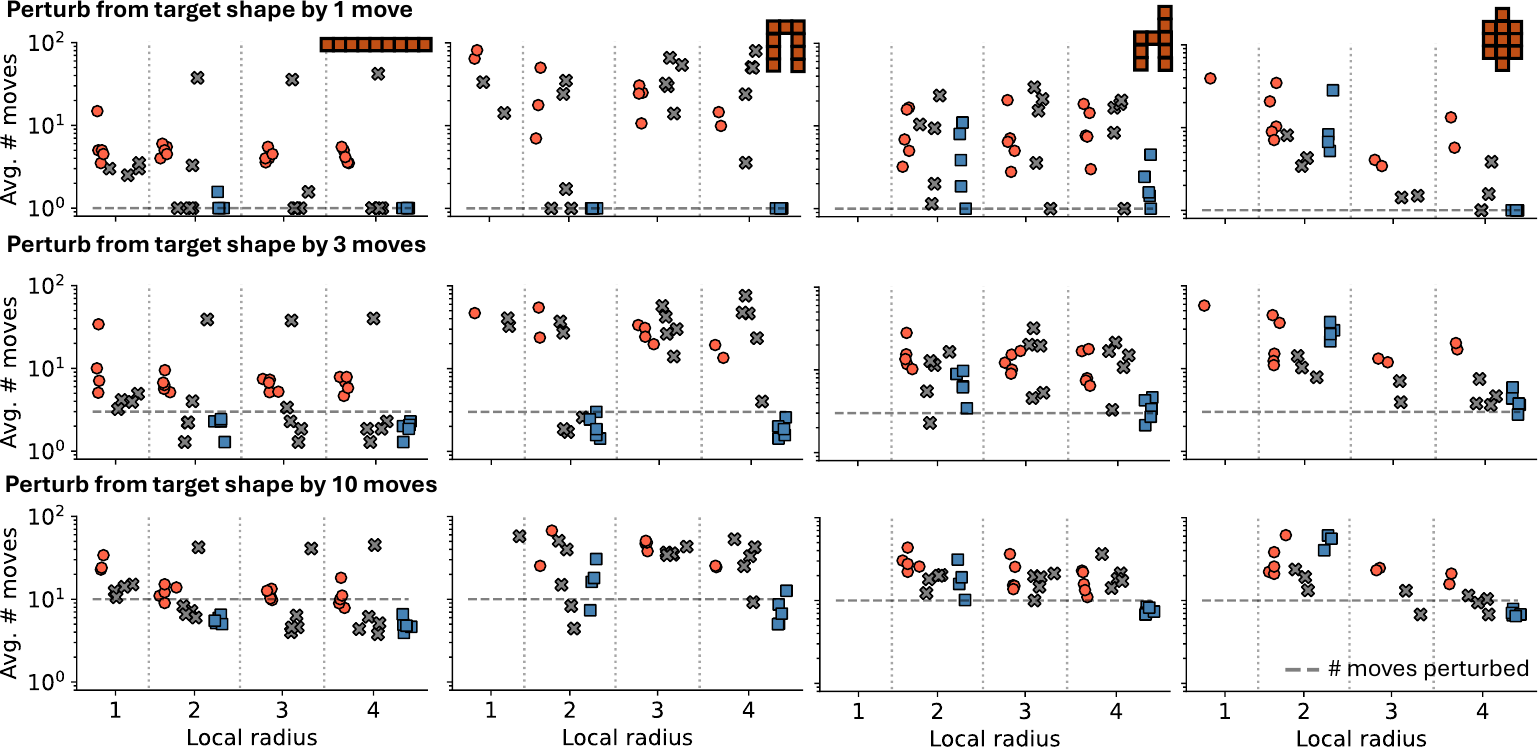}
    \caption{Average number of moves required for the networks in Fig.~\ref{fig:training} to reach their target shape when initialised in the target shape perturbed by (top) one, (middle) three, and (bottom) ten random moves.
    We only show results for networks that achieved a success rate above $99$\% on this task.}
    \label{fig:repairing}
\end{figure}

We explore the effect of locality by varying the size of the local neighbourhood, determined both by the kernel size and the number of convolutions. 
For each shape, six models are trained: MR-CNN with one, two, three, and four $3\times 3$ convolutional layers, and MR-CNN with one and two $5\times 5$ convolutional layers. 
As a reference, we also train models with ordinary $3\times 3$ covolutional layers. 
In the presented results (Fig.~\ref{fig:training}, Fig.~\ref{fig:repairing}), $3\times 3$ MR-CNNs are represented by red circles, $5\times 5$ MR-CNNs by blue squares, and reference CNNs by gray crosses -- all sorted by the effective radius of the convolution operator.
For instance, a model with a single $3\times 3$ convolution is denoted by a local radius of 1, and a model with two such convolutions by a radius of 2.
Similarly, a model with one $5\times 5$ convolution has radius 2, while a model with two $5\times 5$ convolutions has a radius of 4.
Detailed information about the simulation setup can be found in Section~\ref{sec:exps}.

\begin{figure}[t!]
    \centering
    \includegraphics[width=.89\linewidth]{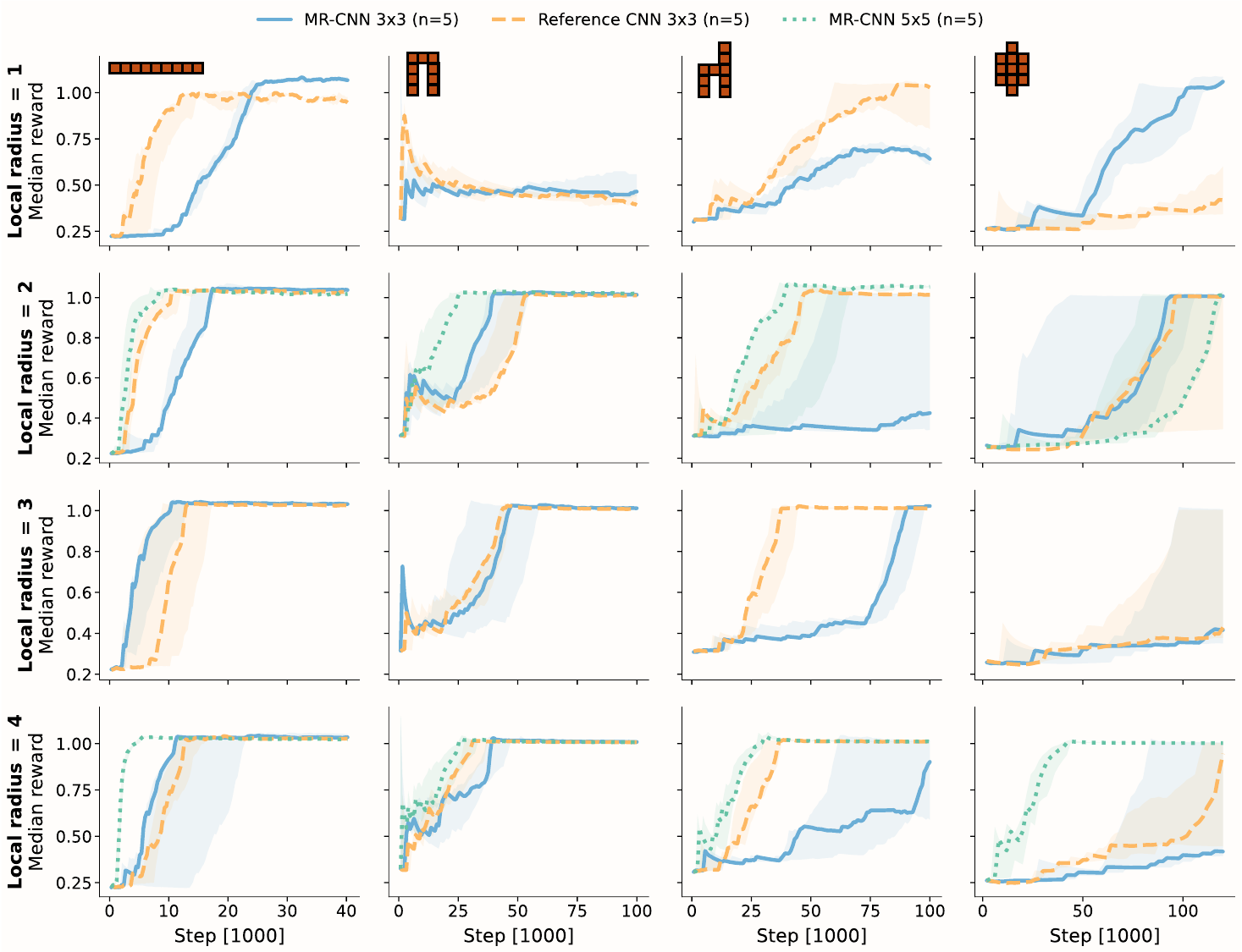}
    \caption{Mean episodic training reward as a function of the number of environment steps (cube moves) during training. Shown is the median of the mean (lines) and first and third quartile (shaded region) for five different seeds.}
    \label{fig:rewards}
\end{figure}
Fig.~\ref{fig:training} shows the training results for all networks on the four target shapes. 
Even for the case of interacting only with the nearest neighbours (local radius equal to 1), the trained networks exhibit a high success rate for all shapes, oftentimes reaching 100\% in at least one trial. 
However, training performance is more stable and consistent as the receptive field increases, which can be seen by reduced variance in the final success rate and required number of moves (e.g., in Fig.~\ref{fig:training}, for $3 \times 3$ MR-CNN on sun-shield, training succeeds only two out of five times).
This is also reflected in the reward curves during training (Fig.~\ref{fig:rewards}), where especially the $3 \times 3$ MR-CNN shows high variance in the training progress.
In all reported experiments, even though action masking is only local, no illegal moves were recorded during test time.

Locality also has a clear effect on the number of moves needed to reach the target shape from a random initialisation: there is a continuous drop in the number of moves with increasing local radius. 
The best performing model is the $5\times 5$ kernel MR-CNN with 2 layers, reaching around 100\% success rate for all shapes and trials, as well as consistently reaching the lowest number of moves. 
It requires on average around $1.89$ moves per cube for reconfiguring into a line, $2.17$ for reconfiguring into a table, $1.43$ for reconfiguring into a chair, and $1.01$ for reconfiguring into a sun-shield. 
In case of the line target shape, we can compare this result with the algorithm introduced in \cite{sung2015reconfiguration}, which always guarantees successful reconfiguration into a line, requiring a number of moves in $\mathcal{O}(n^2)$, where $n$ is the number of cubes.
For comparison, our MR-CNN network with two $5\times 5$ convolutions takes $1.89n$ moves, and the 1-layer $3 \times 3$ reference CNN requires $3.32n$ moves (measured for $n=9$).
However, different from \cite{sung2015reconfiguration}, we do not report scaling behaviour here, and a proper comparison will require simulations for larger cube ensembles in the future.
Moreover, in \cite{zhang2021reinforcement}, reconfiguration trajectories between any two configurations that scale in $\mathcal{O}(n)$ are reported, but for the opposite scenario studied in our work: every cube is unique.

The functional benefits of the inductive biases are less clear, as most improvements tend to originate from the increased local radius.
In fact, the two best performing architectures are the reference CNN as well as the $5\times 5$ MR-CNN network which captures both rotation and mirror symmetries.
In contrast, the $3\times 3$ MR-CNN performs worse than the reference CNN, most likely because it cannot capture the mirror symmetry.

\subsection{Efficiency and stability close to the target shape}

Next, we investigate how efficiently the learned networks reconfigure structures towards the target shape -- addressing how stable the ensemble dynamics is close to the target shape.
For that aim, we evaluate the networks at test-time in the following setup: given a target state, we apply $m$ legal random moves, and let the network self-correct the ensemble back to the target shape. 
Once again we measure success rate and number of moves required, for $m = \{1, 3, 10\}$. 
This test is performed for all target shapes and all trained models. 
Results in terms of number of moves are presented in Fig.~\ref{fig:repairing}, where all trials with success rate $<1\%$ are filtered out. 
As we can see, some of our networks, for local radius $\geq2$, perform within the expected bounds: they find an optimal path back to the target configuration with a number of moves close to the number of perturbations $m$ (given by the dashed line) -- or in case of larger $m$, they find an even shorter path back to the target shape.
Overall, the $5\times 5$ MR-CNN architecture performs best on this benchmark, consistently producing good results for all shapes and local radii.

\subsection{Reconfiguring between target shapes by switching neural network parameters}

\begin{figure}[b!]
    \centering
    \includegraphics[width=.9\linewidth]{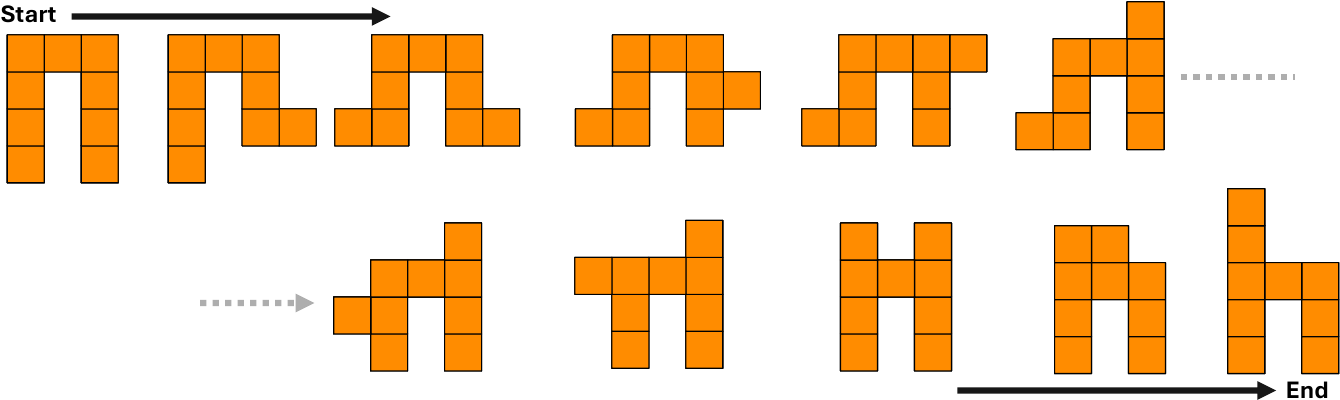}
    \caption{Reconfiguring between different target shapes. Here, we initialise the network in the table shape and let it reconfigure to a chair.}
    \label{fig:assemble1}
\end{figure}

\begin{figure}[b!]
    \centering
    \includegraphics[width=\linewidth]{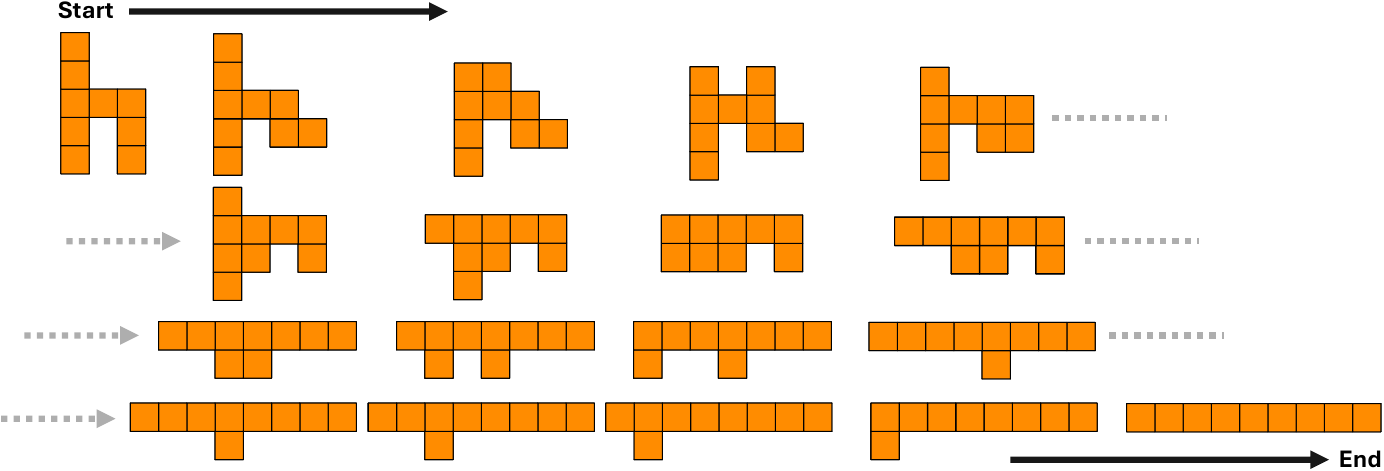}
    \caption{Reconfiguring between different target shapes. Here, we initialise the network in the chair shape and let it reconfigure to a line.}
    \label{fig:assemble2}
\end{figure}

Lastly, since the models learned to reconfigure into their target shapes from any initial state, this means we can morph the ensemble between target shapes by swapping out the network parameters. 
This is illustrated in Fig.~\ref{fig:assemble1}, where we initialise the ensemble in the table configuration and apply the $5\times 5$ 2-layer MR-CNN model trained on chairs to reconfigure the ensemble.
Subsequently, in Fig.~\ref{fig:assemble2} we let the network reconfigure further by loading the network parameters of a model trained to assemble into the line shape.

\section{Discussion}

We examined the feasibility of decentralised control of pivoting cube ensembles, focusing on (i) the effect of locally constrained information passing between cubes, and (ii) the potential of including inductive biases in the neural network architectures controlling the cubes.

In general, we found that reconfiguration is possible even in case of nearest neighbour interactions (i.e., $3 \times 3$ kernels), although more global information exchange has a strong positive effect on efficiency.
In our experiments, this is achieved by stacking several layers of $3 \times 3$ convolutions, leading to improved results while keeping individual interactions local.
However, in practice, having several convolutional layers means regularly exchanging information between cubes at the same time, which requires a global clock signal.
Nevertheless, this is a rather simple global signal that can be made easily available throughout the ensemble.
Moreover, methods utilising asynchronous message passing between cubes can be explored in future work to further reduce the dependence on global signals.

For low local radius, in some cases individual models performed well, but the variation over several training runs was high. Thus, to improve training, model distillation could be applied: training a more global model and consequently transferring its knowledge to a localised one in a supervised training mode. 

We use ensembles of limited size in our experiments. 
This is partly due to issues in scaling observed when going to larger ensembles, stemming from the exponentially growing search space which prevents the model to ever reach the target configuration during training. 
One possible solution to this challenge is curriculum learning, a machine learning strategy where during training, task difficulty is gradually increased \cite{wang2021survey}.
In our case, this translates as follows: in the beginning of training, the initial configuration of the ensemble is only a few steps away from the target shape. 
Incrementally, the initialisation is moved further away from the target by increasing the number of steps.
In the end, after an initial period of growing task difficulty, initial configurations are generated randomly -- as was the case without curriculum learning. 
Initial experimentation with this strategy showed promise, although problems remained with choosing when to jump from the curriculum to the full task difficulty, which sometimes led to training suddenly failing. 
Identifying solutions for larger ensembles, both with the help of additional strategies and more computational power, is definitely a direction worth pursuing in future work.

In this work, we clearly differentiate between the training and testing phase. We treat training as ``before deployment'' of the ensemble, where access to the global state of the ensemble is still available. Thus, during training, we allow for full connectivity checks of the ensemble to mask actions, while during testing, this check is only done locally. To properly evaluate how well the ensembles reconfigure, we do check for connectivity and completion of the target configuration during testing, but without affecting the reconfiguration process. This choice of masking was made to speed up training, although it is an open question whether training with local masking is already sufficient.

Ideally we would like to perform asynchronous moving of multiple cubes in parallel, avoiding the ``global voting'' of actions currently implemented. 
However, this requires a method for ``inhibiting'' cubes from moving to avoid crashing. 
For example, we could have ``local voting'', where cubes within a certain local radius from each other vote which cube is allowed to move.
The winning cube is then free to move, while the others are inhibited from moving until the move of the winner is executed, ensuring that cubes do not crash into each other or disconnect the ensemble. 
Similarly, in future work, the checks performed to identify whether a move is legal could be performed by the control network itself. 
As we already use methods that respect locality to perform these checks, we are confident that this task can be offloaded to the neural networks.

Overall, globality seems to have a much larger impact on performance than inductive bias.
Hence, a promising candidate architecture for further studies is the $3 \times 3$ CNN without rotation invariant kernels, as it combines good performance with local information aggregation that can be stacked to allow cubes to assess more global information about the ensemble.
Nevertheless, the inductive biases also lead to a reduction in the number of model parameters with mostly no degradation in performance, which is beneficial for onboard applications where memory is sparse.
In future work, instead of applying inductive biases, an attention mechanism similar to graph attention \cite{velivckovic2017graph} could be explored, which allows a weighing of the information coming from surrounding cubes.

\section{Conclusion}

This work takes a successful step towards identifying algorithms that enable the decentralised self-assembly of cube ensembles, showing efficient and effective reconfiguration into a number of shapes with local interactions only. 
The main contribution is the demonstration that repeated local exchange of information between cubes is sufficient to enable reconfiguration.
In particular, we found that neural network architectures with inductive biases provide only a minor advantage to using standard neural networks -- although they offer a reduction in required memory while performing similarly to standard neural architectures.
Although we used CNNs here to mimic local information aggregation, in a real hardware system, the operation would be more akin to message passing along the graph of the ensemble.

We are confident that our approach can be scaled up to larger ensembles in future work -- especially since it only encompasses standard neural networks that take feature vectors of neighbouring cubes as input -- and our implementation is easily transferable to 3D. We further believe the proposed method can also be mapped to other systems such as sliding cubes, cube ensembles with mechanical alignment features (e.g., for docking) \cite{team2024reconfiguration}, and CubeSat swarms that can connect to form large-scale space infrastructure.

\subsection*{Acknowledgements}
\label{acknowledgements}
D.D. was funded by the Horizon Europe's Marie Sklodowska-Curie Actions (MSCA) Project 101103062 (BASE).
N.D. acknowledges support through the ESA Student Internship Programme.
Y. Q. is supported by the project Dutch Brain Interface Initiative (DBI2) with project number 024.005.022 of the research programme Gravitation which is (partly) financed by the Dutch Research Council (NWO).
Calculations were performed using supercomputer resources provided by the Austrian Scientific Cluster (ASC), and the High Performance Computing (HPC) Cluster at Donders Centre for Cognitive Neuroimaging in Nijmegen. 

\subsection*{Declaration of competing interest}

The authors have no competing interests to declare that are relevant to
the content of this article.

\printbibliography
\addcontentsline{toc}{section}{References}

\end{document}